\pdfoutput=1
\documentclass[11pt]{article}

\usepackage[]{acl}

\usepackage{times}
\usepackage{latexsym}
\usepackage{multirow}
\usepackage{graphicx}
\usepackage{makecell}
\usepackage{color,soul}
\usepackage{enumitem}
\usepackage{amsmath}

\usepackage[T1]{fontenc}
\usepackage[utf8]{inputenc}

\usepackage{microtype}

\title{Conversational QA Dataset Generation with Answer Revision}

\author{Seonjeong Hwang \\
  Graduate School of Artificial Intelligence \\
  POSTECH, Pohang, South Korea \\
  \texttt{seonjeongh@postech.ac.kr} \\\And
  Gary Geunbae Lee\thanks{\,\,\,Corresponding author} \\
  Computer Science and Engineering \\
  Graduate School of Artificial Intelligence \\
  POSTECH, Pohang, South Korea \\
  \texttt{gblee@postech.ac.kr} \\}

\begin{document}
\maketitle
\begin{abstract}

Conversational question--answer generation is a task that automatically generates a large-scale conversational question answering dataset based on input passages.
In this paper, we introduce a novel framework that extracts question-worthy phrases from a passage and then generates corresponding questions considering previous conversations.
In particular, our framework revises the extracted answers after generating questions so that answers exactly match paired questions.
Experimental results show that our simple answer revision approach leads to significant improvement in the quality of synthetic data. 
Moreover, we prove that our framework can be effectively utilized for domain adaptation of conversational question answering.

\end{abstract}

\section{Introduction}

Conversational question answering (CQA) involves answering questions by considering a given text as well as previous conversations. 
To facilitate research on CQA, a range of datasets have been proposed in recent years \cite{choi2018quac,reddy2019coqa,campos2020doqa,anantha2020open,adlakha2022topiocqa}. 
However, building a robust CQA system for a specific domain requires a large-scale domain-specific dataset; moreover, obtaining such a dataset is considerably expensive and time-consuming.

To resolve this issue, in our previous study, we had proposed a conversational question--answer generation (CQAG) framework that automatically creates multiturn question--answer (Q--A) pairs from given passages \cite{hwang2021study}. 
The framework is a two-stage architecture that adopts contextual answer extraction (CAE) and conversational question generation (CQG). 
Considering previous conversations, the CAE module extracts the next question-worthy phrase from the passage, and then the CQG module generates the conversational question corresponding to the phrase.
However, the framework has the limitation that the error may propagate to the question generation stage and even to data generation for subsequent turns if improper answers are extracted by the CAE module.

In this paper, we introduce a CQAG framework with answer revision (CQAG-AR), in which the conversational question generation with answer revision (CQG-AR) module revises the extracted answer to a more suitable one immediately after generating a question.
In experiments, we synthesize CQA data using CQAG-AR and then evaluate CQA models trained on these synthetic data. 
Results reveal that answer revision by the CQG-AR module leads to absolute improvements of up to 13.4\% and 15.3\% in F1 score and exact match (EM), respectively, for the CQA models. 
Furthermore, fine-tuning the Wikipedia-domain CQA model on different synthetic data increases EM by up to 13.1\%, showing that our framework is beneficial for CQA domain adaptation.

\begin{figure*}[t]
\centering
\includegraphics[width=13cm]{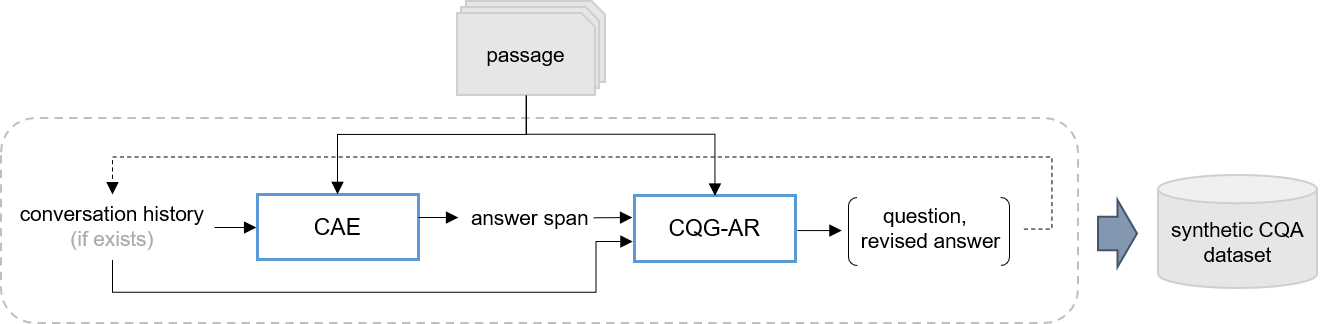}
\caption{\label{tab:main_architecture} Overview of CQAG-AR. Synthetic Q--A pairs are used as conversation history to generate the subsequent Q--A pairs (dotted line).}
\end{figure*}

\section{Related Work}

CQG aims to create conversational questions based on input text. 
It can be subdivided into answer-aware and answer-unaware approaches. 
Answer-aware CQG generates conversational questions corresponding to prepared answers \cite{gao2019interconnected}. 
\citet{gu2021chaincqg} exploited accumulated representations of previous conversations to generate the current question by successively encoding answers and questions that constitute conversation history.
By contrast, answer-unaware CQG synthesizes conversational questions without given answers \cite{wang2018learning,pan2019reinforced,qi2020stay}. 
Further, \citet{nakanishi2019towards} introduced a framework that first finds the location of points of interest in the passage, identifies question types, and subsequently generates conversational questions.

CQAG attempts to automatically construct CQA data for various domains.
In our previous study \cite{hwang2021study}, we designed a vanilla CQAG that generated multiturn Q--A pairs based on a given passage in an autoregressive manner. 
However, the framework has a drawback in that the validity of the extracted answer directly affects the quality of the conversation.

\section{Methods}

Figure \ref{tab:main_architecture} illustrates a CQAG-AR generation pipeline.
To generate a question $q_t$ and answer $a_t$ for the $t$-th turn of conversations, our framework obtains a passage $p$ and conversation history $h_t = ((q_1, a_1), ... , (q_{t-1}, a_{t-1}))$ as inputs. 
The CAE module extracts a probable answer span $a^s_t$ considering these inputs.
Next, the CQG-AR module generates a conversational question $q_t$ and revised answer $a^r_t$ given the inputs and the extracted answer span $a^s_t$.
Finally, we use the revised answer $a^r_t$ as the answer $a_t$.
The modules do not employ the conversation history to synthesize the Q--A pair for the first turn of conversations.
$t$ is omitted in all notations in the following description.

\subsection{Contextual Answer Extraction}

From a given passage $p$, the CAE module extracts an answer span $a^s$ that is most likely to be of interest to a questioner, considering the conversation history $h$, i.e., $P(a^s \mid p, h)$.
We implemented the module using a pretrained BERT \cite{devlin2018bert} with two fully connected (FC) layers at the top \cite{hwang2021study}.  
Each FC layer predicts the index of start and end tokens of the potential answer span in the passage:
\begin{equation*}
\begin{gathered}
prob^s_i = \mathrm{Softmax}(\mathrm{FC}^s(\mathrm{BERT}(p, h)))[i], \\
prob^e_j = \mathrm{Softmax}(\mathrm{FC}^e(\mathrm{BERT}(p, h)))[j],
\end{gathered}
\end{equation*}
where $prob^s_i$ ($prob^e_j$) represents the probability for the $i$-th ($j$-th) token in the passage being the start (end) token of the answer span.

During generation, the top $k$ answer candidates whose start and end indices are $i$ and $j$ are extracted according to the sum of probabilities $prob^s_i + prob^e_j$.
The CAE module outputs the answer span with the highest sum of probabilities after deduplicating the candidates compared with the answers used in the conversation history.
If the candidate set is empty after deduplication, generation is terminated.
To train the module, we computed the sum of cross-entropy losses between predicted start and end indices and the ground truth indices.

\subsection{Conversational Question Generation with Answer Revision}

Considering the input passage and conversation history, the CQG-AR module generates a conversational question and then revises the answer span that is extracted by the CAE module.
The module first considers that the extracted answer span is proper for use as an answer and modifies it if it is inappropriate.
To enable this process, we collected training examples of passage $p$, conversation history $h$, answer span $a^s$, and revised answer $a^r$, which contained positive (\emph{proper} $a^s$, $a^r$) and negative (\emph{improper} $a^s$, $a^r$) pairs.
The module can preserve proper answers extracted by the CAE module by learning positive examples. 
Additionally, negative examples teach the module how to correct improper answer spans with better answers to the generated questions.
We devised two negative sampling techniques to collect improper answer spans from proper ones.

\subsubsection{\label{sec: generating sample} Generating Negative Samples}

For the positive examples, we set ground truth answers of the CQA dataset (e.g., QuAC \cite{choi2018quac}) to both \emph{proper} $a^s$ and $a^r$.
The main experiments were conducted with CoQA \cite{reddy2019coqa}, which contains free-form answers paired with rationales extracted from passages. 
To obtain proper answer spans from CoQA, answer spans with the highest F1 score compared to the free-form answer from the rationale were extracted. 
The \emph{improper} $a^s$ was obtained from the \emph{proper} $a^s$ by using the following techniques.

\noindent \textbf{Answer Span Expansion}  
If the extracted answer contains several key phrases, it may be unsuitable as an answer for a single question. 
In addition, unnecessary words around the answer span should be discarded if they are extracted together. 
To cover these cases, we generated the \emph{improper} $a^s$ by additionally connecting surrounding words of random length to the front or the rear of the \emph{proper} $a^s$. 
However, if the sample was extended to phrases that were answers of other Q--A pairs, the model could confuse the target $a^r$. 
Therefore, we ensured that the sample did not overlap with answers for other turns.

\noindent \textbf{Answer Span Reduction} 
Some important words that constitute a complete answer may be omitted when extracting the answer span.
This phenomenon may risk creating invalid Q--A pairs. 
To create these types of \emph{improper} $a^s$, we removed a random number of words from both ends of the \emph{proper} $a^s$. 
Examples of negative sampling are included in Appendix \ref{sec:negative sampling}.

\subsubsection{Modeling}
When the passage $p$, conversation history $h$, and answer span $a^s$ are given, the CQG-AR module sequentially generates the conversational question $q$ for the input answer span $a^s$ and then revises the answer span $a^s$ based on the generated question $q$:
\begin{equation*}
\label{tab:CQG-AR equation}
\begin{gathered}
P(q, a^r|p, h, a^s) = \prod_{i=1}^{<q>} P(q_i|p, h, a^s, q_{1:i-1}) \\ \,\,\,\,\,\,\,\,\,\,\,\,\,\,\,\,\,\,\,\,\,\,\,\,\,\,\,\,\,\,\,\, \times \prod_{j=1}^{<a^r>} P(a_j^r|p, h, a^s, q, a_{1:j-1}^r),
\end{gathered}
\end{equation*}
where $a^r$ denotes the revised answer and $<\cdot>$ indicates the length of the element.

We implemented the CQG-AR module using T5 \cite{raffel2019exploring}. 
We focused on the masked self-attention mechanism of Transformer \cite{vaswani2017attention}, where the decoder utilizes knowledge of previously generated tokens to predict the current token. 
To revise the answer in a form that is more natural to the question, the module outputs the modified answer immediately after the question is generated. 
The answer span was highlighted using a special token so that the content of interest could be effectively recognized in question generation and answer revision. 
To train our module, we computed the cross-entropy loss between the question and answer of the ground truth and the module's prediction.

\section{Experiments}

\begin{table*}[h]
\scriptsize
\renewcommand{\arraystretch}{1.2}
\resizebox{\textwidth}{!}{%
\begin{tabular}{ll|c|cccc}
\hline
\multicolumn{2}{c|}{\multirow{2}{*}{Training data}}                                           & In-domain             & \multicolumn{4}{c}{Out-of-domain}                                                         \\ \cline{3-7} 
\multicolumn{2}{c|}{}                                                                     & Wikipedia             & News                 & Mid/High Sch.        & Literature           & Children's Sto.      \\ \hline
\multicolumn{1}{l|}{\multirow{3}{*}{Synthetic}} & CQAG-Chain                    & 71.3 / 59.6 & 69.2 / 56.9 & 64.1 / 51.4 & 59.4 / 47.6 & 63.1 / 47.6 \\
\multicolumn{1}{l|}{}                           & Vanilla CQAG                       & 71.3 / 59.9           & 67.6 / 55.7          & 65.8 / 52.7          & 60.3 / 48.3          & 66.6 / 50.5          \\
\multicolumn{1}{l|}{}                           & CQAG-AR (ours) & \textbf{83.1 / 73.8}  & \textbf{81.0 / 71.0} & \textbf{74.4 / 63.2} & \textbf{71.7 / 61.0} & \textbf{75.2 / 61.9} \\ \hline
\multicolumn{2}{c|}{Human-annotated}                                                             & 85.8 / 76.4           & 86.3 / 75.9          & 79.0 / 67.6          & 79.0 / 67.8          & 82.5 / 70.1          \\ \hline
\end{tabular}%
}
\caption{\label{tab:CQA results} F1 (\%) and EM (\%) scores of CQA models on the CoQA$^D$ test set for each domain (The highest performances among results for synthetic data are shown in bold.)}
\end{table*}

\subsection{Experimental Setup}

We employed CoQA \cite{reddy2019coqa}, which comprised 8k passages collected from seven different domains and human-annotated conversations.
To investigate whether CQAG-AR could be effectively utilized to construct a CQA system in a new domain, we split the data into \emph{in-domain} (Wikipedia) and \emph{out-of-domain} (children’s stories, literature, news, and middle and high school English exams).
The \emph{in-domain} data were used to train CQAG frameworks, and the quality of synthetic data generated by the trained CQAG frameworks was evaluated using the \emph{out-of-domain} data.

In addition, we used QuAC \cite{choi2018quac} and DoQA \cite{campos2020doqa} to evaluate our framework.
QuAC is based on 13k Wikipedia passages, and DoQA comprises passages collected from FQAs of three practical domains.
Because the other two domains constituted only the test set, we used only the \emph{cooking} domain of DoQA in our experiment.
The CQAG frameworks used in experiments could generate only open-ended types of data.
Therefore, closed-ended (yes and no) and unanswerable types of examples were excluded from the datasets; these were denoted by CoQA$^D$, QuAC$^D$, and DoQA$^D$.

\subsection{Baseline Frameworks}
To evaluate the quality of the synthetic data generated by our framework, we used two baseline CQAG frameworks:

\noindent \textbf{CQAG-Chain} ChainCQG\footnote{The original ChainCQG accepted a rationale-highlighted passage as an input element but we highlighted an answer span in the passage in our experiment. In addition, we implemented the model based on the original source code: \url{https://github.com/searchableai/ChainCQG}.} \cite{gu2021chaincqg} is a state-of-the-art answer-aware CQG model, and CQAG-Chain combines the CAE module of CQAG-AR and ChainCQG as a CQG module.

\noindent \textbf{Vanilla CQAG} \cite{hwang2021study} This framework shares the same CAE module with CQAG-AR but adopts a simple T5-based CQG module. 
The CQG module accepts the same input elements as CQG-AR but generates only conversational questions.

\subsection{CQA with Synthetic Datasets}

In this section, we evaluate synthetic data generated by CQAG-AR and baseline frameworks by conducting the CQA task. 
In the first experiment, CQAG frameworks learned the \emph{in-domain} data of CoQA$^D$ and then generated synthetic CQA data based on the passages extracted from \emph{in-domain} and \emph{out-of-domain} data. 
Note that we constructed synthetic training and validation sets from original splits of CoQA. 
We implemented a simple CQA model using T5 \cite{raffel2019exploring} and trained several CQA models using human-annotated data (CoQA$^D$) and synthetic datasets. The training details, an example of synthetic conversations, and statistics of the synthetic data can be found in Appendix \ref{sec:details} and \ref{sec:synthetic data}.

Table \ref{tab:CQA results} presents F1 and EM scores of CQA models, which learned the synthetic data, on the test set\footnote{\url{https://github.com/google/BIG-bench}} of CoQA$^D$. 
The CQA models derived from CQAG-AR outperformed other models, showing significant margins of 11.8\% and 13.9\% for in-domain data, and average margins of 10.5\% and 12.5\% for out-of-domain data in F1 and EM, respectively, compared with those derived from the vanilla CQAG. 
These results demonstrate that the answer revision approach is considerably beneficial in terms of generating valid CQA datasets.
However, we additionally found that the out-of-domain CQA models showed lower performances than the in-domain models across all CQAG frameworks. 
This result implies that CQAG frameworks are less robust when extracting valid CQA data from out-of-domain passages.

\begin{table}[h]
\centering
\small
\renewcommand{\arraystretch}{1.2}
\begin{tabular}{ccc}
\hline
Training data                                                                  & \#Training examples & F1 \\ \hline
Human-annotated                                                                           & 3.7k                  & 45.1     \\ \hline
\multirow{2}{*}{\begin{tabular}[c]{@{}c@{}}Synthetic\\ (CQAG-AR)\end{tabular}} & 3.7k                  & 51.5     \\
                                                                               & 4.7k                 & 53.1     \\ \hline
\end{tabular}%
\caption{\label{tab:DoQA} CQA performances on the test set of DoQA$^D$.}
\end{table}

In Table \ref{tab:DoQA}, we compare the evaluation results on the test set of DoQA$^D$ (cooking) for CQA models trained on human-annotated data (DoQA$^D$) and synthetic data. 
We obtained the synthetic data by training CQAG-AR using QuAC$^D$ and then generating the data from the passages of DoQA training and validation sets. 
According to the results, the CQA model trained on synthetic data, which has the same number of examples with the human-annotated data, significantly outperformed the model trained on human-annotated data. 
Moreover, our framework generated a larger number of examples than the ones in the original DoQA$^D$, which improved the F1 score of the CQA model by 1.6\%.
In particular, examples in QuAC$^D$, which were used to train CQAG-AR, were irrelevant to the cooking domain. 
This result indicates that our framework effectively creates synthetic CQA data for an unseen cooking domain.

\subsection{Human Study}

\begin{table*}[h!]
\renewcommand{\arraystretch}{1.2}
\resizebox{\textwidth}{!}{%
\begin{tabular}{l|cl|r}
\hline
Revision type     & Passage                                                                                                                                                                                                            & \multicolumn{1}{c|}{Q-A}                                                                                      & Percentage \\ \hline
Preservation      & \begin{tabular}[c]{@{}c@{}}... It covers and has a population of \hl{2.06 million}. It is a parliamentary\\ republic and a member of the United Nations, European Union, and NATO. ...\end{tabular}                 & \begin{tabular}[c]{@{}l@{}}Q: How many people live in Slovenia?\\ A: 2.06 million\end{tabular}                & 65.2\%     \\ \hline
Reduction         & \begin{tabular}[c]{@{}c@{}}Buckinghamshire (or), abbreviated \hl{Bucks, is a county in South East} \\ \hl{England} which borders Greater London to the south east, Berkshire to the ...\end{tabular}                     & \begin{tabular}[c]{@{}l@{}}Q: Where is it located?\\ A: South East England\end{tabular}                       & 15.3\%     \\ \hline
Expansion         & \begin{tabular}[c]{@{}c@{}}... The group hired Frederick G. Kilgour, a former \hl{Yale University} \\ \hl{medical school librarian}, to design the shared cataloging system. ...\end{tabular}                            & \begin{tabular}[c]{@{}l@{}}Q: Who was he?\\ A: a former Yale University medical school librarian\end{tabular} & 14.2\%     \\ \hline
Multiple revision & \begin{tabular}[c]{@{}c@{}}... Discogs currently contains over 8 million releases, by nearly 4.9 million\\ artists, across over \hl{1 million labels, contributed from nearly 346,000} contributor ...\end{tabular} & \begin{tabular}[c]{@{}l@{}}Q: And how many labels?\\ A: over 1 million\end{tabular}                           & 2.0\%      \\ \hline
Complete change   &  \begin{tabular}[c]{@{}c@{}}... Selective breeding for fast growth, egg-laying ability, conformation, plumage \\ and docility took place \hl{over the centuries}, and modern breeds ...\end{tabular} & \begin{tabular}[c]{@{}l@{}}Q: How did breeds change over time?\\ A: selective breeding\end{tabular}  & 3.4\%      \\ \hline
\end{tabular}%
}
\caption{\label{tab:revision distribution}Distribution of the answer-revision types in the CQG-AR module. (The answer spans extracted by the CAE module are highlighted.)}
\end{table*}

In Table \ref{tab:revision distribution}, we classify the synthetic examples according to revision types. 
The distribution shows that 65.2\% of answer spans extracted by the CAE module were preserved without any modification, while the other 34.8\% of answers were revised.
This demonstrates that the CQG-AR module could recognize invalid answer spans and selectively modify the answers. 
Furthermore, we found that the module could perform more complex revisions such as “multiple revision” and “complete change” though the CQG-AR module learned only examples for "reduction" and "expansion" obtained by negative sampling.

Further, we conducted human evaluation to compare the quality of synthetic data generated by the vanilla CQAG and our CQAG-AR.
From the two synthetic datasets presented in Table \ref{tab:CQA results}, 120 examples (30 examples from each out-of-domain dataset) were sampled and three volunteers were asked to rate 80 examples according to the criteria listed in Appendix \ref{sec:assessment items}.

\begin{table}[h]
\centering
\scriptsize
\renewcommand{\arraystretch}{1.2}
\begin{tabular}{llrr}
\hline
                                                                                 &                   & \multicolumn{1}{c}{Vanilla CQAG} & \multicolumn{1}{c}{CQAG-AR} \\ \hline
\multirow{3}{*}{\begin{tabular}[c]{@{}l@{}}Question\\ Connectivity\end{tabular}} & Dependent         & 67.9\%                     & 66.7\%                      \\
                                                                                 & Independent       & 27.7\%                     & 30.6\%                      \\
                                                                                 & Unnatural         & 4.5\%                      & 2.8\%                       \\ \hline
\multirow{3}{*}{\begin{tabular}[c]{@{}l@{}}Answer\\ Correctness\end{tabular}}    & Correct           & 64.2\%                     & 85.8\%                      \\
                                                                                 & Partially correct & 23.3\%                     & 4.2\%                       \\
                                                                                 & Incorrect         & 12.5\%                     & 10.0\%                      \\ \hline
\end{tabular}%
\caption{\label{tab:assessment results} Results of human evaluation for synthetic data generated by vanilla CQAG and CQAG-AR.}
\end{table}

Although 2.9\% more synthetic questions of CQAG-AR are independent of the previous conversations than the questions of the vanilla CQAG in Table \ref{tab:assessment results}, the synthetic questions of the two frameworks show almost similar evaluation results.
These results prove that the vanilla CQAG, which performs only question generation, and CQAG-AR, which performs question generation and answer revision in an end-to-end manner, can generate questions of similar quality.
However, 21.6\% more synthetic answers of CQAG-AR were judged as correct answers compared with those of the vanilla CQAG.
Furthermore, the number of partially correct answers was considerably reduced through answer revision.
This reveals that answer revision is effective in correcting inappropriate answer spans extracted by the CAE module into correct answers that match well with the question.

\subsection{\label{sec:domain_adaptation_exp} Domain Adaptation}

In this experiment, we tested the effectiveness of the synthetic data generated by CQAG-AR in adapting the CQA model from the Wikipedia domain to new domains (\emph{out-of-domain}).
We trained CQA models, which were initialized with parameters of T5-Large, in three training settings. 
In the \textbf{In-Man} setting, we trained the CQA model on the Wikipedia data of CoQA$^D$. 
In the \textbf{Out-Syn} setting, CQA models learned out-of-domain synthetic data. 
Finally, the model of In-Man setting was fine-tuned with synthetic data of each out-of-domain case in the \textbf{In-Man $\rightarrow$ Out-Syn} setting.

\begin{figure}[]
\centering
\includegraphics[width=7cm]{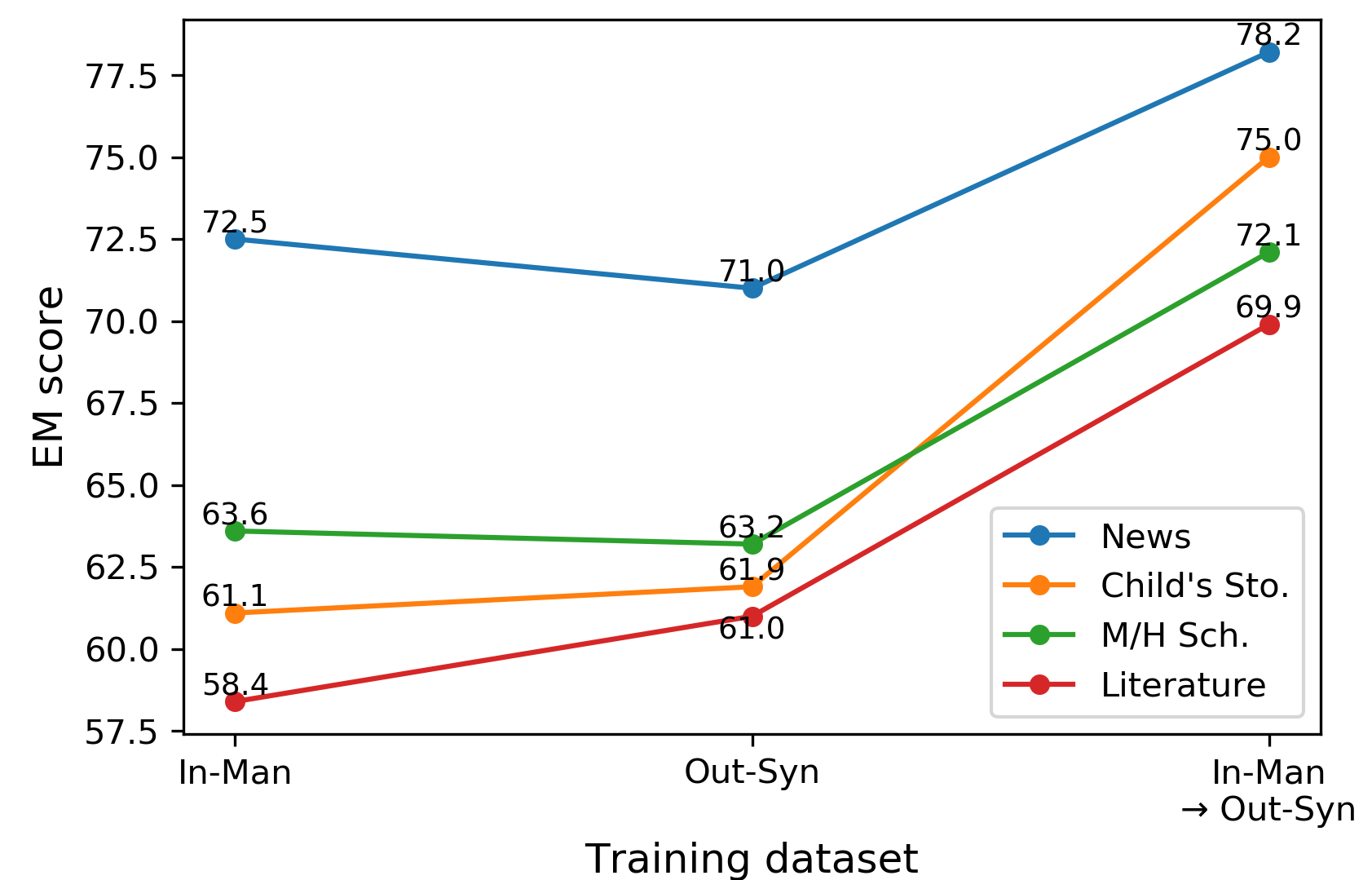}
\caption{\label{tab:domain adaptation} EM scores of several CQA models on the CoQA$^D$ test set for each domain.}
\end{figure}

As shown in Figure \ref{tab:domain adaptation}, the models in the Out-Syn setting yielded results similar to those of the model in the In-Man setting while exhibiting better EM scores in the two domains.
Notably, fine-tuning the Wikipedia CQA model using our synthetic data (In-Man $\rightarrow$ Out-Syn) improved the EM scores of the model by an average of 9.7\% across all domains. 
This result indicates that our framework can be effectively utilized for domain adaptation in CQA.

\section{Conclusion}
In this paper, we propose CQAG-AR, which automatically synthesizes high-quality CQA data. 
Our framework comprises CAE and CQG-AR modules. 
Considering the conversation history, the CAE module extracts a question-worthy phrase from a given passage, and then the CQG-AR module generates a conversational question while revising the extracted answer to make it more suitable. 
Experimental results show that CQAG-AR outperforms baseline frameworks in terms of generating high-quality CQA data. 
In addition, fine-tuning a Wikipedia-domain CQA system on our synthetic data for out-of-domain cases improves the model performances by significant margins.

\section*{Acknowledgements}
We would like to thank Professor Yunsu Kim (POSTECH) for his expert advice. This work was supported by SAMSUNG Research, Samsung Electronics Co.,Ltd., and also supported by the MSIT(Ministry of Science and ICT), Korea, under the ITRC(Information Technology Research Center) support program(IITP-2022-2020-0-01789) supervised by the IITP(Institute for Information \& Communications Technology Planning \& Evaluation)

\bibliography{custom}

\newpage

\appendix
\onecolumn

\section{\label{sec:negative sampling} Examples of Negative Sampling}

\begin{figure}[h]
\centering
\includegraphics[width=9cm]{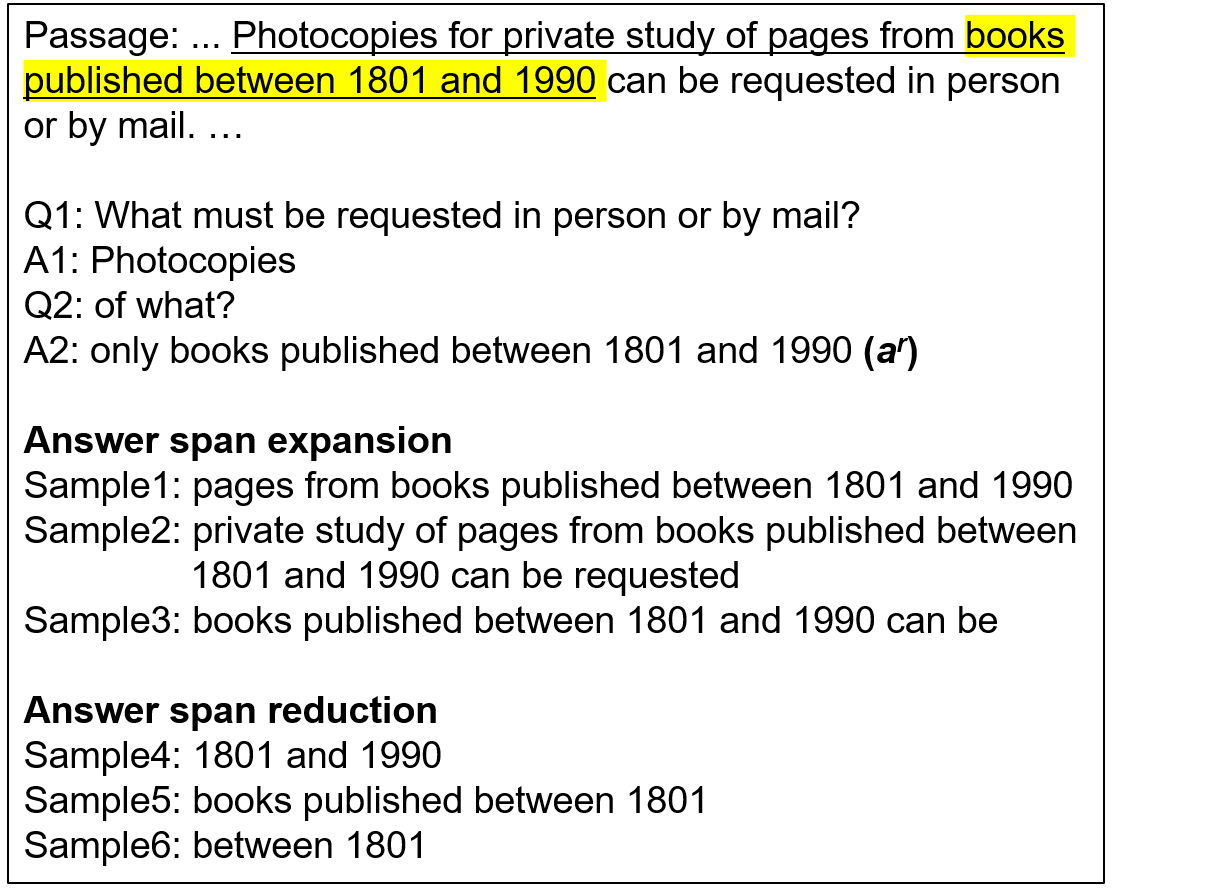}
\caption{Example of negative sampling applied to one passage in CoQA. The rationale for A2 is underlined, and \emph{proper} $a^s$ corresponding to A2 is highlighted in the passage. A2 is $a^r$ for both the \emph{proper} and \emph{improper} $a^s$ samples.}
\end{figure}

\section{\label{sec:details}Training Details}
To implement the CAE module, we used parameters of BERT-large-uncased. Only the previous two pairs of Q--A were used as the conversation history to extract the $i$-th answer span and passages $p$ longer than 512 tokens were truncated with a stride of 128 tokens:
\begin{align*}
    & \textbf{Input:}\,\,\,\mathrm{[CLS]}\,\,q_{i-2}\,\,a_{i-2}\,\,q_{i-1}\,\,a_{i-1}\,\,\mathrm{[SEP]}\,\,\mathrm{truncated\,}p\,\,\mathrm{[SEP]}
\end{align*}
For the CQG-AR module, we initialized the module with parameters of T5-large. The input and target sequences for generation of the $i$-th Q--A pair are as follows:
\begin{align*}
    & \textbf{Input:}\,\,\,a^s_i\mathrm{\,highlighted}\,p\,\mathrm{[SEP]\,[A]}\,a_{i-4}\,\mathrm{[Q]}\,q_{i-4}\,\,...\,\,\mathrm{[A]}\,a_{i-1}\,\mathrm{[Q]}\,q_{i-1}\mathrm{[A]}\,a^s_i \\
    & \textbf{Target:}\,\,\,\mathrm{[Q]}\,q_i\,\mathrm{[A]}\,a^r_i\,\mathrm{[EOS]}
\end{align*}
Because the input sequence length of the T5 encoder was limited, we truncated the passage $p$ at the 32nd token after the location of $a^s_i$. 
Regarding the conversation history, only the previous four Q--A pairs were used. 
Special tokens ([Q] and [A]) were added before each question and answer, and [A] and $a^s_i$ were appended to the end of the input sequence. 
[Q] was used as a bos token, and answer generation started when [A] was returned after predicting the question.

We utilized the Transformers library and pre-trained parameters from HuggingFace\footnote{\url{https://huggingface.co/}} and conducted experiments using A100 GPUs. 
Further, AdamW was used as the optimization algorithm with a batch size of 4 and a learning rate of 3e-5. 
In addition, a learning rate scheduling algorithm was applied and the warm-up period was set to the initial 10\% of the total steps. 
For CAE, we optimize the module based on F1 score between predicted answer span and the ground truth.
For CQG-AR, beam search with a beam size of 4 was used during data generation. 
The best module was selected based on METEOR \cite{banerjee2005meteor} and BERTScore \cite{zhang2019bertscore} on the synthetic development set.

For CQA, we designed a simple T5-based model that accepted the concatenation of the passage, conversation history, and question as input and then generated the answer to the input question. 
We initialized our model with T5-Large and trained the model with AdamW, setting the batch size between 4 and 8 and the learning rate to 3e-5. 
When fine-tuning the Wikipedia-domain CQA model with synthetic data in Section \ref{sec:domain_adaptation_exp}, we fine-tuned the model for one epoch, with a batch size of 1 and the learning rate between 1e-7 and 1e-6. 
We employed the same training and decoding strategies used for the CQG-AR module.

\section{\label{sec:synthetic data}Synthetic Data}
\subsection{\label{sec:data example}Example of Synthetic Conversation}

\begin{table}[h]
\renewcommand{\arraystretch}{1.2}
\scriptsize
\resizebox{\textwidth}{!}{%
\begin{tabular}{|p{12cm}|}
\hline
\textbf{Passage}:  CHAPTER IV. \hl{Signor Andrea D'Arbino}, searching vainly through the various rooms in the \hl{palace} for Count Fabio d'Ascoli, and trying as a last resource, \hl{the corridor leading to the ballroom and grand staircase}, discovered his friend \hl{lying on the floor in a swoon}, without any living creature near him. Determining to avoid alarming the guests, if possible, D'Arbino first sought help in the \hl{antechamber}. He found there \hl{the marquis's valet}, assisting \hl{the Cavaliere Finello} (who was just taking his departure) to put on his cloak. While Finello and his friend carried Fabio to an open window in the antechamber, the valet procured \hl{some iced water}. This simple remedy, and the change of atmosphere, proved enough to restore the fainting man to his senses, but hardly--as it seemed to his friends--to his former self. They noticed a change to \hl{blankness and stillness} in his face, and when he spoke, an indescribable alteration in the tone of his voice. "I found you in a room in the corridor," said D'Arbino. "What made you faint? Don't you remember? Was it the heat?" Fabio waited for a moment, painfully collecting his ideas. He looked at the valet, and Finello signed to the man to withdraw. "Was it the \hl{heat}?" repeated D'Arbino. "No," answered Fabio, in strangely hushed, steady tones. "I have seen the face that was behind the \hl{yellow mask." "Well?" "It was the face of my dead wife}." "Your dead wife!" "When the mask was removed I saw \hl{her face}. Not as I remember it in the pride of her youth and beauty--not even as I remember her on her sick-bed--but as I remember her in her \hl{coffin}." \\
                                                               \\
\begin{tabular}[c]{@{}l@{}}
\textbf{Conversation} \\
Q1: Who was searching for Fabio d'Ascoli?\\ A1: Signor Andrea D'Arbino\\ 
Q2: Where was he searching?\\ A2: the palace\\ 
Q3: What was his last resort?\\ A3: the corridor leading to the ballroom and grand staircase \\
Q4: What did he find?\\ A4: lying on the floor in a swoon\\
Q5: Where did he seek help first?\\ A5: the antechamber\\
Q6: Who helped him?\\ A6: the marquis's valet\\
Q7: Who helped him put on his cloak?\\ A7: the Cavaliere Finello\\
Q8: What did the valet give him?\\ A8: some iced water\\
Q9: What change did his friends notice?\\ A9: a change to blankness and stillness in his face\\
Q10: What did D'Arbino say was the cause?\\ A10: heat\\
Q11: What was behind the mask?\\ A11: the face of my dead wife\\
Q12: What did I see?\\ A12: her face\\
Q13: How did I remember her?\\ A13: in her coffin\\
\end{tabular} \\ \hline
\end{tabular}%
}
\caption{Samples of generated Q-A pairs using CQAG-AR from a Wikipedia passage in CoQA. Answer spans before revision are highlighted in the passage in order.}
\end{table}

\newpage

\subsection{\label{sec:data info}Statistics of Synthetic Data}

\begin{table}[h]
\centering
\renewcommand{\arraystretch}{1.2}
\resizebox{0.45\columnwidth}{!}{%
\begin{tabular}{l|cc}
\hline
                    & Synthetic dataset & CoQA \\ \hline
\#Words in question & 5.6               & 5.4  \\
\#Words in answer   & 3.0               & 2.6  \\
\#Turns per passage & 12.1              & 15.1 \\ \hline
\end{tabular}%
}
\caption{Average number of words in the questions and answers, and the average number of conversation turns in CoQA and our synthetic data extracted from CoQA passages.}
\end{table}

\begin{table}[h]
\centering
\renewcommand{\arraystretch}{1.2}
\resizebox{0.45\columnwidth}{!}{%
\begin{tabular}{l|rr|rr}
\hline
\multirow{2}{*}{Domain} & \multicolumn{2}{c|}{\#Passages}                      & \multicolumn{2}{c}{\#Q--A pairs}                      \\
                  & \multicolumn{1}{c}{Train} & \multicolumn{1}{c|}{Dev} & \multicolumn{1}{c}{Train} & \multicolumn{1}{c}{Dev} \\ \hline
Wikipedia         & 1.6k                      & 0.1k                     & 23.6k                     & 1.5k                    \\
News              & 1.7k                      & 0.1k                     & 21.6k                     & 1.2k                    \\
Mid/High Sch.     & 1.7k                      & 0.1k                     & 22.2k                     & 1.3k                    \\
Literature        & 1.6k                      & 0.1k                     & 17.7k                     & 1.1k                    \\
Children's Sto.   & 0.6k                      & 0.1k                     & 6.3k                      & 1.2k                    \\ \hline
\end{tabular}%
}
\caption{Statistics summarizing the synthetic datasets generated from CoQA passages.}
\end{table}

\section{\label{sec:assessment items} Criteria for Human Evaluation}

\begin{table*}[h]
\centering
\small
\renewcommand{\arraystretch}{1.2}
\begin{tabular}{ll}
\hline
\multicolumn{2}{l}{\textbf{Question Connectivity}} \\ \hline
\multicolumn{1}{l|}{Dependent}          & The current question refers to previous conversations (e.g., via pronoun usage or ellipses).              \\
\multicolumn{1}{l|}{Independent}        & The current question is not dependent on previous conversations.                 \\
\multicolumn{1}{l|}{Unnatural}          & The current question has grammatical errors or overlaps with previous conversations.  \\ \hline
\multicolumn{2}{l}{\textbf{Answer Correctness}}                  \\ \hline
\multicolumn{1}{l|}{Correct}            & Questions are paired with correct answers.                            \\
\multicolumn{1}{l|}{Partially correct}  & Answers are incomplete or contain unnecessary information.              \\
\multicolumn{1}{l|}{Incorrect}          & Not the correct answer to the question.                                 \\ \hline
\end{tabular}%
\caption{Criteria for human evaluation of synthetic CQA data.}
\end{table*}

\end{document}